\pdfoutput=1

\documentclass[11pt]{article}

\usepackage{acl}

\usepackage{times}
\usepackage{latexsym}
\usepackage{comment}
\usepackage{graphicx}

\usepackage[T1]{fontenc}

\usepackage[utf8]{inputenc}

\usepackage{microtype}

\newcommand{\abs}[1]{\vert#1\vert}
\newcommand{\norm}[1]{\vert\vert#1\vert\vert}

\title{Fast Few-shot Debugging for NLU Test Suites}

\author{Christopher Malon \and Kai Li \and Erik Kruus \\
  NEC Laboratories America \\
  4 Independence Way \\
  Princeton, NJ 08540 \\
  \texttt{{malon,kaili,kruus}@nec-labs.com}}

\begin{document}
\maketitle
\begin{abstract}
We study few-shot debugging of transformer based natural language
understanding models, using recently popularized test suites to not just
diagnose but correct a problem.  Given a few debugging examples of a
certain phenomenon, and a held-out test set of the same phenomenon,
we aim to maximize accuracy on the phenomenon at a minimal cost of accuracy
on the original test set.  We examine several methods that are
faster than full epoch retraining.  We introduce a new fast method, which
samples a few in-danger examples from the original training set.
Compared to fast methods using parameter distance constraints or
Kullback-Leibler divergence, we achieve
superior original accuracy for comparable debugging accuracy.
\end{abstract}

\section{Introduction}

When deep transformer models make mistakes, ML engineers have had little
recourse but to collect a better training set and hope the problem is fixed.
Adversarial datasets have exposed a variety of phenomena under which
models trained on common datasets fail, particularly for question answering
and natural language inference \citep{jia-liang-2017-adversarial, gururangan-etal-2018-annotation, kim-etal-2018-teaching, mccoy-etal-2019-right, nie-etal-2020-adversarial, thorne-etal-2019-fever2}.
They have provided new test data to expose problems but not always
new training data to correct them.
Recently, the natural language processing community has adopted methodologies
inspired by software development for probing and testing the capabilities
of a model.
\citet{ribeiro-etal-2020-beyond} introduce CheckList,
which helps users to develop test suites of examples, organized by capability.

Collecting hundreds or thousands of examples for each error phenomenon is
slow, expensive, and not always feasible.  In this paper, we investigate
how just a few examples of a phenomenon (``debugging examples'', which were
not in the original dataset)
can be utilized to correct a model.
The goal is higher accuracy on the phenomenon (``debugging accuracy'')
while retaining accuracy
on the original dataset (``original accuracy'').  This problem differs
from domain adaptation and
few-shot learning because performance must be maintained on original
examples, and no new classes are introduced.

We repurpose published test suites for several natural language understanding
(NLU) tasks as debugging problems,
not just diagnostics.  We identify methods that can update a model using
a few debugging examples without the expense of iterating over the whole
original training set.  We introduce a new fast method that samples
in-danger examples from the original training set to obtain even better
original accuracy for comparable debugging accuracy.


\section{Related work}

Two recent works \citep{zhu, de-cao-etal-2021-editing} study how to modify transformer
language models so that they store updated facts, testing their
approaches on downstream tasks such as zero-shot relation extraction
and closed-book fact checking.  To apply these methods,
one is given a modified fact as an example to train on, and one must
predict the modified fact correctly (success rate) while achieving low
performance deterioration on the original test set.
Because success rate is measured on just one example which is
available at training time, to determine whether the update really
generalizes,
\citet{de-cao-etal-2021-editing} also measures {\em equivalence accuracy}, which reflects
accuracy on paraphrases of the updated fact.

By contrast, our setting provides ten examples (not just one) for a
phenomenon where the predictions are to be updated.
The phenomenon being debugged may involve deeper semantics than
a factoid update, which usually requires only a reassociation of
particular words that appear in the example.
We assume we are given a testing set for the phenomenon, so
we can measure generalization
by directly measuring accuracy on the testing examples
instead of paraphrasing the training examples.

Despite these differences, ideas from these papers provide relevant
ideas that can be used in our debugging setting as well.
One baseline considered by \citet{zhu}, which we call {\em intensive
fine-tuning}, simply takes the updated facts (for us, the debugging
training set) and repeatedly performs gradient descent updates
on them until they are classified correctly.

The proposed approach of \citet{zhu} is to minimize loss on the updated
facts (the debugging set) subject to either an $L^\infty$ or $L^2$ constraint
on the difference of the model parameters.  We consider these as baselines.

As \citet{de-cao-etal-2021-editing} observe, constraining the norm of the parameter update
is only loosely tied to how a parameter change can affect the output of
a model.  For this reason they introduce an approach based on constraining
the Kullback-Leibler divergence between the updated model and the original.
Their proposed method trains a hypernetwork to read a single updated example
and make a change minimizing debugging loss subject to the Kullback-Leibler
divergence constraint.  That does not apply as well to our scenario
of multiple debugging examples, but we borrow the idea of using
Kullback-Leibler divergence to incentivize similar predictions in a
more straightforward baseline.

\citet{sinitsin} introduce a meta-learning method for making a model
that will preserve original accuracy when performing a series of gradient
descent steps to change the label of any particular example.
We are interested in methods that can be applied to any model,
and for real debugging it is not necessary that all examples be
easily relabeled.

Contemporaneously to our work, \citet{pasunuru-etal-2021-continual}
investigate few-shot debugging on error categories that are apparently
too broad to be corrected with just a few examples.
Although they report some success with feature matching methods such as
prototypical networks \citep{snell}, they either
suppose that test examples are identified as needing a correction or not
(i.e. debugging or original), more like domain adaptation, or else
train the prototypical network on a combined training set, which is the
slowness we are trying to avoid.  Our setting requires a single model that can
be applied to all examples without source information.

\section{Method}

\begin{table*}[htb]
\begin{center}
\begin{tabular}{p{1.1in}p{.8in}p{.8in}p{.8in}p{.8in}p{.8in}}
\hline
Test suite & Dog & Or/And & Becoming & People & Passive \\
\hline
Before debugging & (.000, .913) & (.000, .913) & (.002, .913) & (.005, .913) & (.009, .913) \\
\hline
{\em Fast} & & & & & \\
Debug only & (.731, .909) & (1.000, .909) & (1.000, .910) & (.922, .910) & (.819, .910) \\
$L^2$ ($\delta=.1$) & (.704, .909) & (1.000, .909) & (1.000, .911) & (.880, .910) & (.876, .910) \\
$L^\infty$ ($\delta=.1$) & (.704, .909) & (1.000, .909) & (1.000, .911) & (.880, .910) & (.876, .910) \\
K-L ($\lambda=10$) & (1.000, .905) & (1.000, .908) & (1.000, .909) & ( 1.000, .908) & (1.000, .908) \\
Ours & (.731, .909) & (.994, .913) & (1.000, .913) & (.993, .911) & (.975, .912) \\
\hline
{\em Slow} & & & & & \\
Mixed in & (1.000, .913) & (.999, .912) & (1.000, .913) & (.933, .912) & (.859, .912) \\
Oversampling & (1.000, .911) & (1.000, .913) & (1.000, .912) & (.999, .914) & (1.000, .911) \\
\hline
\end{tabular}
\caption{(Debugging accuracy, Original accuracy) on CheckList test suites for
QQP.}
\label{tbl:qqp}
\end{center}
\end{table*}

\begin{table*}[htb]
\begin{center}
\begin{tabular}{p{1.1in}p{1.3in}p{1.3in}p{1.3in}}
\hline
Test suite & Used to but now & Negation with neutral & Opinion matters \\
\hline
Before debugging & (.793, .925) & (.448, .925) & (.616, .925) \\
\hline
{\em Fast} & & & \\
Debug only & (.860, .914) & (1.000, .917) & (.602, .915) \\
$L^2$ ($\delta=.1$) & (.860, .915) & (1.000, .919) & (.600, .915) \\
$L^\infty$ ($\delta=.1$) & (.860, .915) & (1.000, .919) & (.600, .915) \\
K-L ($\lambda=10$) & (.838, .915) & (1.000, .916) & (.538, .920) \\
Ours & (.877, .919) & (1.000, .913) & (.777, .885) \\
\hline
{\em Slow} & & & \\
Mixed in & (.909, .913) & (1.000, .925) & (.673, .923) \\
Oversampling & (.735, .931) & (1.000, .921) & (.512, .928) \\
\hline
\end{tabular}
\caption{(Debugging accuracy, Original accuracy) on CheckList test suites for
SST-2.}
\label{tbl:sst2}
\end{center}
\end{table*}

\begin{table*}[htb]
\begin{center}
\begin{tabular}{p{1.1in}p{.8in}p{.8in}p{.8in}p{.8in}p{.8in}}
\hline
Test suite & After If & P. Participle & Disjunction & Passive & NP/S \\
\hline
Before debugging & (.000, .838) & (.001, .838) & (.005, .838) & (.004, .838) & (.006, .838) \\
\hline
{\em Fast} & & & & & \\
Debug only & (1.000, .813) & (1.000, .804) & (1.000, .807) & (.929, .827) & (1.000, .811) \\
$L^2$ ($\delta=.1$) & (.999, .816) & (.999, .810) & (.999, .812) & (.933, .827) & (.999, .817) \\
$L^\infty$ ($\delta=.1$) & (1.000, .812) & (1.000, .804) & (1.000, .807) & (.933, .827) & (1.000, .811) \\
K-L ($\lambda=10$) & (1.000, .825) & (1.000, .820) & (1.000, .822) & (1.000, .824) & (1.000, .826) \\
Ours & (1.000, .841) & (.926, .835) & (1.000, .836) & (.994, .832) & (.939, .842) \\
\hline
{\em Slow} & & & & & \\
Mixed in & (.468, .835) & (.114, .833) & (.344, .837) & (.791, .835) & (.298, .837) \\
Oversampling & (.920, .836) & (.992, .837) & (1.000, .838) & (.869, .837) & (1.000, .833) \\
\hline
\end{tabular}
\caption{(Debugging accuracy, Original accuracy) on HANS test suites for
MNLI.}
\label{tbl:hans}
\end{center}
\end{table*}

We suppose we are given a model $p_\theta (x,y)$ trained on
training set $X$.  We are also given
debugging training set $X^\prime$, and original
test set $X_{test}$ and debugging test set $X^\prime_{test}$.
These four sets are pairwise disjoint.
We consider the cross-entropy loss
\begin{equation}
\mathcal{L} (x, y; \theta) = - p_\theta (x, y) \log p_\theta (x, y) \ldotp
\end{equation}
Our method initializes $\theta_0 = \theta$ and then performs
{\em intensive fine-tuning} on the debugging set $X^\prime$, by
performing Adam \citep{adam}
iterations $\theta_{t+1} = Adam(\mathcal{L}, X^\prime, \theta_t)$
where $Adam(\mathcal{L}, S, \theta)$ represents the parameter update
achieved by training $\theta$ with respect to the loss $\mathcal{L}$ over
a complete epoch on $S$.
Intensive fine-tuning stops at the minimal step $t = t_{X^\prime}$ such that
$\mbox{argmax}_y p_{\theta_t} (x_i, y) = y_i$ for all
$(x_i, y_i) \in X^\prime$.
We write $\theta_{X^\prime} = \theta_{t_{X^\prime}}$.

Next we collect random samples $W \subset X$ that are misclassified
by $\theta_{X^\prime}$ but not by $\theta$.  In our experiments we select
$\abs{W} = 2 \abs{X^\prime}$ such examples.  Collecting $W$ is a fast
process involving iterating through a random shuffle of $X$ and stopping
when the required number of examples is retrieved.  The expected iteration
time depends only on the error rates and correlation of the errors of the
models and not on the size of the original training set $\abs{X}$.

Finally we restart from the original parameters $\theta$ and intensively
fine-tune using the set $X^\prime \cup W$.  We take $\theta^\prime_0 = \theta$
and iterate Adam
\begin{equation}
\theta^\prime_{t+1} = Adam(\mathcal{L}, X^\prime \cup W, \theta^\prime_t)
\end{equation}
until we reach $t^\prime$ where
$\mbox{argmax}_y p_{\theta^\prime_{t^\prime}} (x_i, y) = y_i$ for all
$(x_i, y_i) \in X^\prime \cup W$.
The resulting $\theta^\prime = \theta^\prime_{t^\prime}$
is the debugged model by our proposed method.


\section{Experiments}

We consider a BERT base model \citep{devlin-etal-2019-bert} implemented in Pytorch
\citep{pytorch} by the HuggingFace Transformers library \citep{wolf-etal-2020-transformers}
for all experiments, with batch size 16 per GPU on 3 or 4 GPU's,
otherwise following default training parameters.

Our data sets are test suites from
HANS \citep{mccoy-etal-2019-right} debugging an MNLI model
\citep{williams-etal-2018-broad} and CheckList \citep{ribeiro-etal-2020-beyond}
debugging models for SST-2 and QQP from GLUE \citep{wang-etal-2018-glue}.
We take test cases with the worst accuracy before debugging, and select 10
examples from each suite for debugging ($X^\prime$) and use the rest
(e.g. 990 examples for HANS) to test debugging ($X^\prime_{test}$).
See the appendix for details.
Our data splits and our code for extracting examples
from CheckList are available for download.\footnote{https://github.com/necla-ml/debug-test-suites}
For HANS we use the BERT cased model and for CheckList we use the uncased model.

\subsection{Fast baselines}

The first of four fast baselines we consider, which is labeled ``debug only,'' performs
intensive fine-tuning on the debugging set $X^\prime$ only, returning the
model $\theta_{X^\prime}$.  In every case we tested, $t_{X^\prime} \leq 3$
epochs over ten examples, so this completed within a minute.

The next baselines from \citet{zhu} are
finding $\theta^\prime$ to minimize $\mathcal{L}(X^\prime, \theta^\prime)$
subject to an $L^\infty$ constraint
${\norm{\theta^\prime - \theta}}_\infty < \delta$ or an $L^2$ constraint
${\norm{\theta^\prime - \theta}}_2 < \delta$.  Following \citet{zhu} we use
$\delta = 0.1$ and implement the optimization as projected gradient descent,
{\em e.g.} for $L^\infty$, taking a gradient descent step
from $\theta_0$ to $\theta$
and projecting the updated parameters back into the $L^{\infty}$ ball as
\begin{equation}
\theta_0 + \min ( \max (\theta - \theta_0, - \delta), \delta)
\end{equation}
limiting the excursion in any coordinate to $\pm \delta$.

The fourth baseline we consider introduces a Kullback-Leibler divergence
on randomly sampled examples from $X$ into the loss:
\begin{equation}
\mathcal{L}^\prime (\theta^\prime) = \mathcal{L}(X^\prime ; \theta^\prime) + \lambda \mathcal{L}_{KL} (X ; \theta^\prime)
\end{equation}
where
\begin{equation}
\mathcal{L}_{KL} (X; \theta^\prime) = \sum_{(x,y) \in X} \sum_{y^\prime}
p_\theta (x, y^\prime) \log \frac{p_\theta (x, y^\prime)}{p_{\theta^\prime} (x, y^\prime)}
\end{equation}
In practice, $\mathcal{L}_{KL} (X; \theta^\prime)$ is estimated on
minibatches from $X$ simultaneously with selecting a minibatch of the same
size from $X^\prime$.

Training on each of these baselines stops when we reach $t^\prime$ where
$\mbox{argmax}_y p_{\theta^\prime_{t^\prime}} (x_i, y) = y_i$ for all
$(x_i, y_i) \in X^\prime$.  In experiments, this always happens within
three epochs over $X^\prime$ .

\subsection{Slow baselines}

Our first slow baseline is simply to train the model starting with the original
BERT base parameters for three full epochs on randomly shuffled
$X^\prime \cup X$, without accounting for the difference in
size $\abs{X^\prime} << \abs{X}$.  We call this ``mixed in'' training.

Our second baseline (``oversampling'')
equally weights $X^\prime$ and $X$ in the training.
It starts with original BERT base parameters and trains for three full epochs
over $X$, each time taking a batch consisting half of examples from $X$
and half of examples from $X^\prime$, interleaved.  Although the $X$ samples
are sampled without replacement, the $X^\prime$ samples are replaced and
are each seen many times.

\subsection{Results}

We consider the CheckList and HANS test suites for QQP, SST-2, and MNLI
together (Tables~\ref{tbl:qqp},~\ref{tbl:sst2}, and~\ref{tbl:hans}).
Among fast methods, our method has the highest
original accuracy in 11 out of 13 subcases and the highest debugging
accuracy in 6 out of 13.  This makes it a better choice for retaining
original accuracy out of several fast, good methods for improving debugging
accuracy.
Kullback-Leibler divergence, which ranks first most often among fast methods 
in debugging accuracy, only ranks first in original accuracy once out of
13 subcases.  Notably, both methods frequently outperform the debug only
approach in debugging accuracy, showing that
sampling non-debugging examples helps achieve an update
that generalizes better even on the debugging phenomenon.

Considering slow methods, oversampling achieves maximal debugging
accuracy on 8 of 13 subcases and best original accuracy on 8 of 13.
On HANS, mixing the debugging examples into the full training set is not
sufficient for them to be learned, though this method achieves
reasonable debugging accuracy on the other datasets.

\begin{figure}[htb]
\includegraphics[width=3in,height=2.2in]{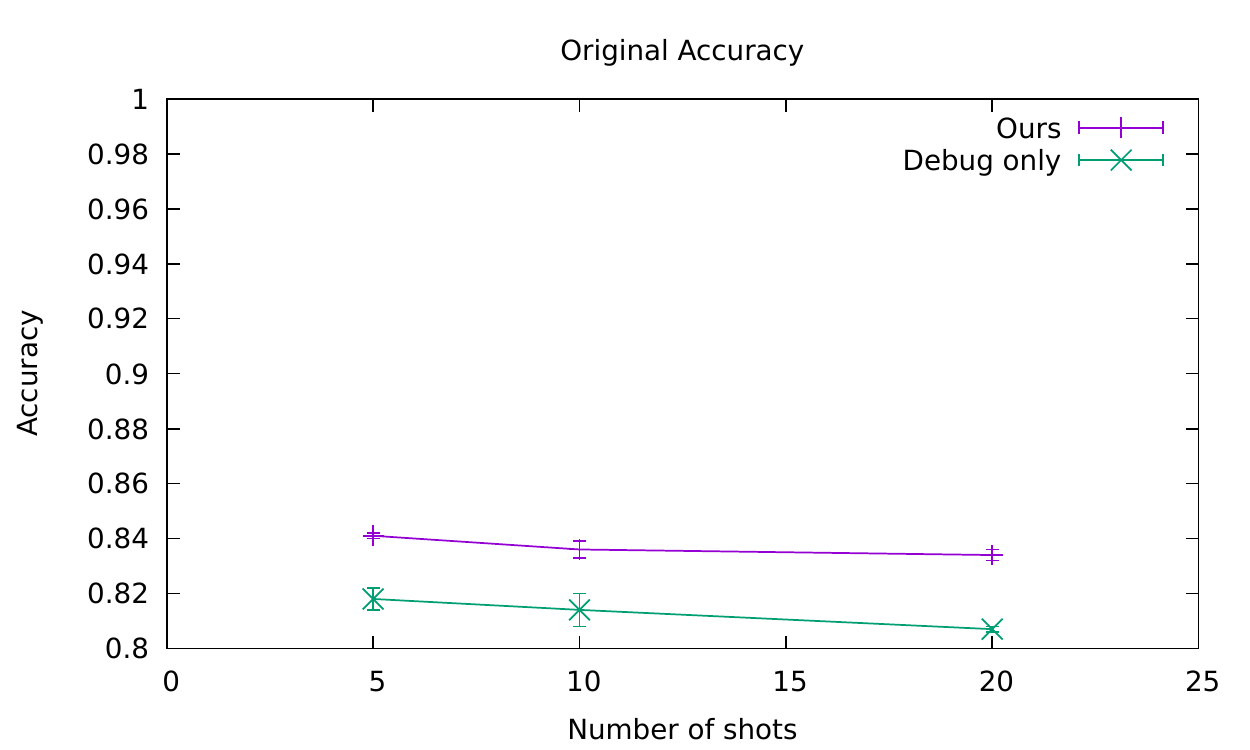} 
\includegraphics[width=3in,height=2.2in]{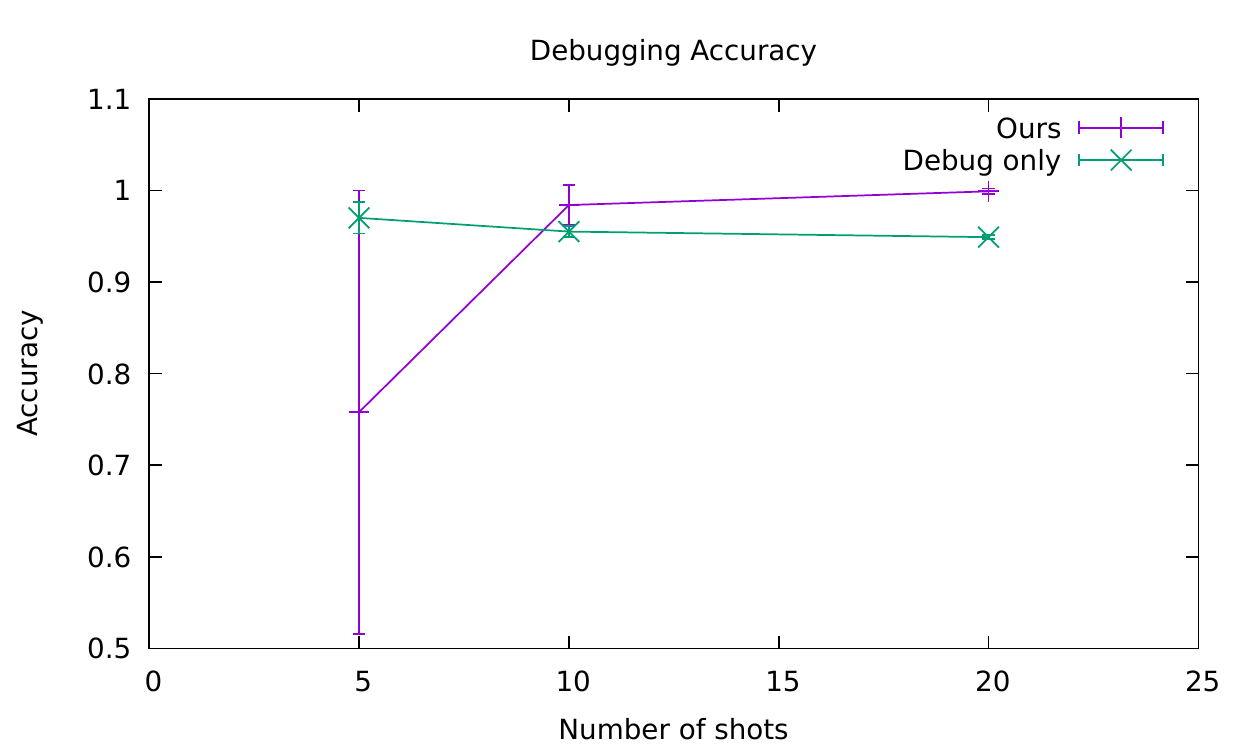}
\caption{Comparing
our method to debug-only intensive fine-tuning for
different numbers of shots.}
\label{fig:bothfig}
\end{figure}

{\bf Number of shots and stability.}
Besides the 10 shot setting described above, we compare our method to
``debug only'' intensive fine-tuning for 5 shots and 20 shots.
Results are shown for HANS's \verb+cn_after_if_clause+ test suite
in Figure~\ref{fig:bothfig}.  Each experiment is repeated, sampling
eight different sets of debugging and in-danger examples.
The standard deviation in accuracy over the samples is indicated by
the error bars around each mean result in the figure.

Five shots is too few to be sure of good debugging accuracy.
Our method achieves significantly higher debugging accuracy and original
accuracy, compared to intensive fine-tuning, with ten or twenty shots.
With twenty shots the debug only method loses original accuracy, possibly
due to the tightened constraints of classifying more debugging examples
correctly.

{\bf Other base models.}  We repeat 10-shot experiments using Electra
\citep{electra}
instead of BERT.  Using Electra, our method has the highest original
accuracy among fast methods in 7 out of 13 subcases and the highest
debugging accuracy in 8 out of 13.

\begin{table}[htb]
\begin{center}
\begin{tabular}{lr}
\hline
Method & Seconds \\
\hline
{\em Fast} & \\
Debug only & 10.89 \\
$L^2$ & 14.74 \\
$L^\infty$ & 15.85 \\
K-L & 14.79 \\
Ours - {\em total} & 25.29 \\
\, {\em debug-only fine-tuning} & 10.89 \\
\, {\em finding new misclassifications $W$} & 2.86 \\
\, {\em final fine-tuning} & 11.54 \\
\hline
{\em Slow} & \\
Mixed in & 12663.14 \\
Oversampling ({\em estimated}) & 25326.28 \\
\hline
\end{tabular}
\caption{Model debugging time in seconds.}
\label{tbl:time}
\end{center}
\end{table}

{\bf Time.}  
Intensive fine-tuning usually finishes after a few small batches,
but collecting the 20 misclassified examples potentially can require
more evaluations.
On QQP these can be found in 1/60 of an epoch (forward only)
and at worst (on ``negation with neutral'' of SST-2) in 1/5,
yielding roughly 720x and 60x speedups over oversampling (three
epochs, forward and back, alternating with debugging examples), respectively.

In Table~\ref{tbl:time} we collect total timings for each debugging procedure
on HANS's \verb+cn_after_if_clause+ test suite,
including the time our method needs to collect the new misclassifications $W$
from the original MNLI training set.  Whereas the slow methods require hours
to update the model, all the fast methods finish in a matter of seconds.

\section{Conclusion}

We study the new problem of few-shot debugging natural language
understanding problems on narrowly defined test suites,
addressing a real-life need not addressed by past benchmark datasets.
Intensive fine-tuning on debugging examples with a few newly misclassified
examples is substantially faster than full epoch retraining, and retains
superior accuracy on the original dataset in more of our tests than any other
fast method, for competitive debugging accuracy.  Kullback-Leibler
regularization may achieve better debugging accuracy, but its original
accuracy is lagging, probably because it samples randomly rather
than focusing on the
newly misclassified examples that the debugging examples are opposed to.
Our results suggest a way for practitioners to quickly address problems
in deployed systems and inspire the search for more refined ways of
using debugging information.

To further this research, there is a need for test suites that are not
constructed by templates, so that the debugging phenomena are less easily
learned, and yet not too broad to be taught in the few-shot setting.
This limitation forced us to focus on relatively small differences
in accuracy.  Because our method requires only a few debugging examples,
it should be practical to construct test suites by hand or by manually
organizing existing misclassifications.

\bibliography{anthology,debugging}
\bibliographystyle{acl_natbib}

\clearpage

\appendix

\section{Test suites}
\label{sec:appendix}

{\bf HANS}.  The Multi-Genre Natural Language Inference (MNLI)
dataset \citep{williams-etal-2018-broad} tests natural language inference
in multiple domains, such as fiction, letters, telephone speech,
and government reports.  It is framed as a three-class classification
problem of pairs of sentences, as entailment, neutral, or contradiction.
MNLI provides matched and mismatched development and test sets, in which
the mismatched setting tests domains not present in the training data.
Here we consider a model trained on MNLI and take its accuracy
on the matched development set as a measure of its original performance.

HANS \citep{mccoy-etal-2019-right} is a dataset that compiles phenomena that may not be
adequately learned from the MNLI training set.  Three heuristics
(lexical overlap, sequence, or constituent) for generating challenging
examples are considered, each with ten subcases, for a total
of thirty subcases.  Templates are used to generate one thousand
training and one thousand test examples for each.
For our experiments,
we individually consider the five subcases on which the MNLI model attains
the lowest accuracy before debugging.
Since we are interested
in few-shot debugging, we randomly take ten of
the HANS training examples for a subcase as our debugging set $X^\prime$ but
use the rest (990) as $X^\prime_{test}$ for testing debugging performance.

HANS examples are labeled only as entailment or non-entailment, without
specifying whether the non-entailments should be contradiction or neutral
classifications.  When training on a non-entailment example, we
backpropogate through a logit representing the total non-entailment
probability specified by the three-class model
\begin{eqnarray}
p_\theta (x, nonent) & = & p_\theta (x, n) + p_\theta (x, c) \\
\log p_\theta(x, nonent) & = & \log \frac{e^{l_n} + e^{l_c}}{e^{l_e} + e^{l_n} + e^{l_c}}
\end{eqnarray}
where $l_y = \log e^{p_\theta (x, y)}$ and $y$ ranges over the entailment
($e$), neutral ($n$), and contradiction ($c$) classes.

{\bf CheckList}.  CheckList \citep{ribeiro-etal-2020-beyond} compiles test suites
for sentiment analysis (SST-2) and duplicate question detection (QQP),
two datasets which can be found in the GLUE benchmark \citep{wang-etal-2018-glue}.

SST-2 binarizes classifications from Stanford Sentiment Treebank \citep{socher-etal-2013-recursive}
into positive or negative, but some test suites of CheckList utilize
a neutral target label.  We eliminate such test suites.  Some test
suites of CheckList test invariance or directional properties of
classifications ({\em e.g.} whether two examples are classified with the same
label, without specifying what that label should be) and we eliminate those
as well, focusing only on suites with given labels for each example.
We are left with three suites on which accuracy of the base SST-2 model
before debugging is worse than the overall SST-2 accuracy.

Quora Question Pairs (QQP) is already a binary classification task
and no adjustments to the test suites are needed.  Again, we consider
only test suites consisting of individually labeled examples.
We take the five suites where the base QQP model achieves lowest
accuracy before debugging.  For each suite, we randomly pick 10 examples
for $X^\prime$ and put the rest (usually about 1000) in $X^\prime_{test}$.

The full names of the tests utilized are as follows.

For HANS: \verb+cn_after_if_clause+, \verb+sn_past_participle+,
\verb+cn_disjunction+, \verb+ln_passive+, and \verb+sn_NP/S+.

For SST-2: Used to but now, Hard negation of positive with neutral stuff
in the middle should be negative, and My opinion is what matters.

For QQP: Do you have to X your dog before Y it, A or B is not the
same as A and B, What was person's life before becoming X / What was person's
life after becoming X, Traditional SRL wrong active passive swap,
and Traditional SRL wrong active passive swap with people.

\end{document}